\documentclass[conference]{IEEEtran}
\IEEEoverridecommandlockouts
\usepackage{amsmath,amssymb,amsfonts}
\usepackage{algorithmic}
\usepackage[numbers]{natbib}
\usepackage{graphicx}
\usepackage{textcomp}
\usepackage{multirow}
\usepackage[final]{hyperref} 
\hypersetup{
	colorlinks=true,       
	linkcolor=blue,        
	citecolor=blue,        
	filecolor=magenta,     
	urlcolor=blue         
}
\usepackage{xcolor}
\def\BibTeX{{\rm B\kern-.05em{\sc i\kern-.025em b}\kern-.08em
    T\kern-.1667em\lower.7ex\hbox{E}\kern-.125emX}}
\begin{document}

\title{Next-Generation License Plate Detection and Recognition System using YOLOv8\\
}

\author{\IEEEauthorblockN{Arslan Amin}
\IEEEauthorblockA{School of Electrical Engineering\\ and Computer Science (SEECS), \\National University of Sciences\\ and Technology (NUST), \\ Islamabad, 44000, Pakistan\\
Email:aamin.mscs19seecs@seecs.edu.pk}
\and
\IEEEauthorblockN{Rafia Mumtaz}
\IEEEauthorblockA{School of Electrical Engineering\\ and Computer Science (SEECS), \\National University of Sciences\\ and Technology (NUST), \\ Islamabad, 44000, Pakistan\\
Email:rafia.mumtaz@seecs.edu.pk}
\and
\IEEEauthorblockN{Muhammad Jawad Bashir}
\IEEEauthorblockA{School of Electrical Engineering\\ and Computer Science (SEECS), \\National University of Sciences\\ and Technology (NUST), \\ Islamabad, 44000, Pakistan\\
Email:mbashir.msds20seecs@seecs.edu.pk}
\and
\IEEEauthorblockN{Syed Mohammad Hassan Zaidi}
\IEEEauthorblockA{Ghulam Ishaq Khan Institute of Engineering Sciences \\and Technology (GIKI), Topi, District, Swabi, \\Khyber Pakhtunkhwa 23460, Pakistan,\\
Email:prorector\_acad@giki.edu.pk}}

\maketitle

\begin{abstract}
In the evolving landscape of traffic management and vehicle surveillance, efficient license plate detection and recognition are indispensable. Historically, many methodologies have tackled this challenge, but consistent real-time accuracy, especially in diverse environments, remains elusive. This study examines the performance of YOLOv8 variants on License Plate Recognition (LPR) and Character Recognition tasks, crucial for advancing Intelligent Transportation Systems. Two distinct datasets were employed for training and evaluation, yielding notable findings. The YOLOv8 Nano variant demonstrated a precision of 0.964 and mAP50 of 0.918 on the LPR task, while the YOLOv8 Small variant exhibited a precision of 0.92 and mAP50 of 0.91 on the Character Recognition task. A custom method for character sequencing was introduced, effectively sequencing the detected characters based on their x-axis positions. An optimized pipeline, utilizing YOLOv8 Nano for LPR and YOLOv8 Small for Character Recognition, is proposed. This configuration not only maintains computational efficiency but also ensures high accuracy, establishing a robust foundation for future real-world deployments on edge devices within Intelligent Transportation Systems. This effort marks a significant stride towards the development of smarter and more efficient urban infrastructures.

\end{abstract}


In the intricate web of modern urban infrastructure, the capability to identify vehicles swiftly and accurately has emerged as a cornerstone for numerous applications. From traffic management and congestion control to security surveillance and automated toll collection, the importance of efficient license plate detection and recognition has grown exponentially \cite{Tippannavar2023}.

Traditional methods of license plate detection, while commendable in their efforts, often grapple with challenges such as variability in environmental conditions, diverse license plate designs across regions, and the need for real-time processing. The dynamic nature of these challenges has fueled a relentless pursuit of more adaptive and robust solutions.

The field of License Plate Recognition (LPR) has witnessed a significant evolution, pivoting from the foundational tasks of detection, extraction, and recognition of license plates to embracing more nuanced challenges such as deblurring, denoising, and geometric transformations to bolster recognition accuracy \cite{Khan2023}. Initially, the exploration in this domain was channeled through employing neural networks to tackle individual facets of the problem. However, a paradigm shift occurred as subsequent efforts aimed at orchestrating neural networks to oversee the entire process, with proposals extending to utilizing up to five networks for discerning the presence of a license plate within an image. Various neural network architectures such as Convolutional Neural Networks (CNNs), Generative Adversarial Networks (GANs), and Recurrent Neural Networks (RNNs) have been utilized in License Plate Recognition (LPR). CNNs excel in character detection and recognition, GANs can generate synthetic training data, and RNNs may aid in recognizing character sequences on license plates. A notable trajectory in this field has been the strategic modification of the YOLO \cite{YOLO} network to accelerate processing speed, reflecting a concerted effort to match the real-time exigencies of surveillance and traffic monitoring applications. 

Deep learning, particularly in the domain of CNNs, has shown immense promise in revolutionizing object detection tasks, including license plate recognition. Among the various architectures and models that have gained prominence, the YOLO series stands out due to its ability to detect objects in real-time with remarkable accuracy. Moreover, the ingenuity embedded in methods like N-YOLO \cite{NYOLO}, which deviates from traditional image size reduction to employing fixed-size image patches, and innovative approaches like CornerNet \cite{CornerNet} and HoughNet \cite{HoughNet}, that shift the gaze from top-bottom to bottom-up object detection strategies, underscores the unrelenting quest for better LPR performance, especially in recognizing smaller objects like license plates amidst a larger image frame. The latest in this lineage, YOLOv8, brings with it refinements and optimizations that potentially make it a frontrunner for tasks requiring swift and precise detections.

In this work, we investigate the potential of state-of-the-art YOLOv8 for license plate detection and recognition by training its three variants—medium, small, and nano—to identify the most apt model for different scenarios. Our key contributions are:
\begin{itemize}
\item Curated LPR and Character Recognition datasets.
\item Showcased the effectiveness of YOLOv8 models in License Plate and Character Recognition tasks, with the 'small' variant attaining a precision of 0.945 and 0.92 on LPR and Character Recognition test data respectively.
\item Introduced a simple character sequencing method, with 99.8\% accuracy in accurately sequencing detected characters.
\end{itemize}

\section{Literature Review}

The domain of vehicle license plate detection and character recognition continues to be a focal point of study due to its pivotal role in modern Intelligent Transportation Systems (ITS). The intricacies of this domain arise from a myriad of variables such as fluctuating weather conditions, diverse illumination scenarios, and inherent data acquisition noises that are encountered in real-world scenarios. Further complexities are introduced by the necessity for real-time performance, a critical attribute for applications in state-of-the-art ITS. Among the multitude of approaches aimed at addressing these challenges, a noteworthy endeavor is encapsulated in the presented paper, which delineates a unified application for License Plates Detection (LPD) and Character Recognition (CR), tailored for Iranian vehicle license plates. This demographic of license plates presents a unique set of challenges including varying resolution and layouts, a limited array of digits/characters, assorted background colors, and disparate font sizes.

One of the proposed license plate detection and character recognition algorithms based on a combined feature extraction model and BPNN is effective and efficient under various complex backgrounds, with a recognition accuracy of 97.7\% and a consuming time of 46.1ms \cite{Xie2018}. \cite{Tourani2020A} leveraged a distinctively fine-tuned version of the You Only Look Once (YOLO) V3 platform for both phases, managing to extract Persian characters in two automated steps. Through extensive training and testing on a comprehensive dataset of 5719 images, it achieved an end-to-end accuracy of 95.05\%, and demonstrated real-time performance with an average processing time of 119.73 milliseconds for real-time data, without necessitating any pre-processing, calibration, or configuration procedures. \cite{Shi2023} addressed license plate recognition challenges, utilizing enhanced YOLOv5 and a GRU + CTC network, achieving 98.98\% recognition precision, outperforming conventional algorithms, and exhibiting robustness in complex environments. \cite{Ammar2023} put forward a multi-stage system tackled real-time vehicle and Saudi license plate identification challenges using integrated detectors, classifiers, and trackers, optimizing edge GPU performance. It enhanced recognition accuracy by 13\% and 40\% for car models and license plates respectively, achieving 17.1 FPS on a Jetson Xavier AGX device. \cite{Padmasiri2022} unveiled a hardware-efficient license plate recognition solution for low-resource edge devices, adapting to diverse illumination with novel neural networks. Despite resource constraints, it showcased competitive performance against server-grade solutions in real-time, day and night operations.
Employing a CNN for character recognition, \cite{Kaur2022} proposed ALPR system, bolstered by pre-processing and morphological operations, adeptly managed diverse license plate types and nighttime operations. It achieved an overall accuracy of 98.13\%, marking a competent solution for varied vehicle types and conditions, while addressing the challenges with unusual license plate varieties and nighttime functionality faced by conventional systems. \cite{Jun2023} examined ViT's efficacy in license plate corner prediction, juxtaposed against ResNet and MobileNet models. Despite ViT's size-related performance, ResNet exhibited superior absolute performance. The analysis of 600 ViT backbone models highlighted the challenges and intricacies of transitioning from CNN to ViT, emphasizing ViT's complex input image handling and training process. Another study by \cite{Xia2022} explored YOLOv7's proficiency in detecting license plates in complex traffic scenarios, enhancing it with Transformer modules. The modified YOLOv7, exhibiting an AP around 0.88, outperformed original baselines and other YOLO models on the CCPD dataset, showcasing reduced FLOPs and maintained accuracy through various structural modifications. \cite{Alireza2022} introduced a lightweight Visual Transformer-based model for OCR in license plate images, surpassing traditional CRNN and CNN models with a 77.25\% accuracy on a new LicenseNet dataset, while requiring 3.21 times fewer training parameters, demonstrating enhanced efficacy and resource-efficiency on edge devices. Addressing challenges in License Plate Recognition (LPR), \cite{Nascimento2023} showcased a SISR approach employing attention and transformer modules for better LP reconstruction from low-resolution footage. Utilizing synthetic training images, the approach achieved a Structural Similarity Index Measure (SSIM) of less than 0.10, significantly outperforming existing methods in reconstructing low-resolution images both quantitatively and qualitatively. To tackle this with a different approach, \cite{Alhalawani2023} employed a cutting-edge diffusion model for image restoration, outclassing traditional methods like CNNs and GANs. Trained on a Saudi license plate dataset, the method displayed a notable improvement of 12.55\% and 37.32\% in PSNR over SwinIR and ESRGAN respectively, alongside a 4.89\% and 17.66\% enhancement in SSIM. With 92\% of human evaluators favoring the produced images, the research unveils a promising license plate super-resolution solution for surveillance applications.

The reviewed literature highlights significant progress in License Plate Recognition (LPR) and Character Recognition (CR), driven by deep learning. However, challenges persist, particularly in real-time processing on resource-constrained edge devices, handling low-resolution images, and adapting to diverse real-world conditions. Additionally, the need for more comprehensive and diverse datasets for training and testing is evident. Our work aims to address these gaps, proposing a novel approach to enhance accuracy, efficiency, and robustness in LPR and CR systems using state-of-the-art YOLOv8, while fostering better adaptability to varying operational scenarios.

\section{Dataset}

The initial dataset (Table-\ref{tab:datasets}), denoted as the License Plate Reg. No. (LPR) dataset was acquired from Roboflow, encompassing 1,500 annotated images. This dataset is segmented into a training subset with 1,050 images (70\%), a validation subset with 225 images (15\%), and a test subset with 225 images (15\%). Every image within this dataset has been subjected to preprocessing actions such as auto-orientation and resizing to a uniform resolution of 640x640 pixels. This dataset is singular in its focus on the LP Reg. No. class.

Similarly, the Character Recognition dataset (Table-\ref{tab:datasets}), sourced from Roboflow as well, incorporates 25,000 annotated images. This dataset is divided into a training subset with 17,500 images (70\%), a validation subset with 3,750 images (15\%), and a test subset with 3,750 images (15\%). Each image in this dataset has been subjected to preprocessing steps such as auto-orientation and resizing, achieving a standardized resolution of 640x640 pixels. This dataset is diversified into 36 classes, encompassing 26 classes dedicated to alphabet characters from A-Z and an additional 10 classes for numeric characters from 0-9.

\begin{table}[htbp]
\centering
\caption{Datasets for License Plate and Character Recognition}
\begin{tabular}{|c|c|c|c|c|}
\hline
\multirow{2}{*}{\textbf{Dataset}} & \multicolumn{4}{c|}{\textbf{Data Split}} \\
\cline{2-5}
& \textbf{Total} & \textbf{\textit{Train}} & \textbf{\textit{Valid}} & \textbf{\textit{Test}} \\
\hline
\textbf{License Plate Reg. No. (LPR)} & 1,500 & 1,050 & 225 & 225 \\
\hline
\textbf{Characters Recognition} & 25,000 & 17,500 & 3,750 & 3,750 \\
\hline
\end{tabular}
\label{tab:datasets}
\end{table}

\section{Mehthodology}

\begin{figure}[!htb]
    \centering
    \includegraphics[width=9cm]{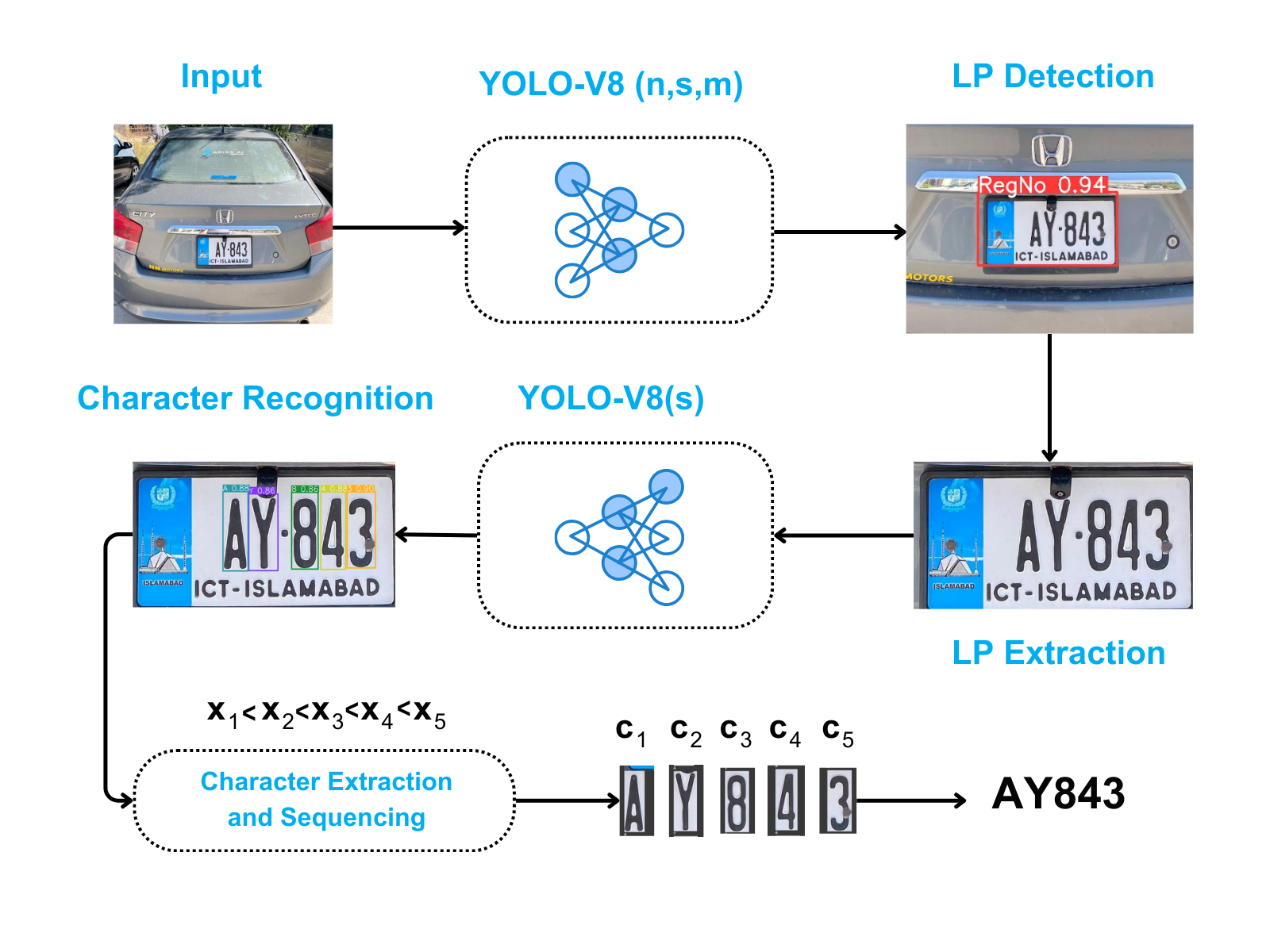}
    \caption{Methodology Overview}
    \label{fig:method}
\end{figure}

This section delineates the procedural steps employed for executing the license plate and character recognition tasks using YOLOv8 across distinct datasets. Figure-\ref{fig:method} illustrates the overview of our methodology.

\subsection{Dataset Preparation}
Two diverse datasets were utilized for this study; one catering to License Plate Recognition (LPR) and the other for Character Recognition tasks. These datasets were meticulously curated and pre-processed to ensure consistency and accuracy in the subsequent training and evaluation phases.

\subsection{YOLOV-8 Architecture}
\begin{figure}[!htb]
    \centering
    \includegraphics[width=4cm]{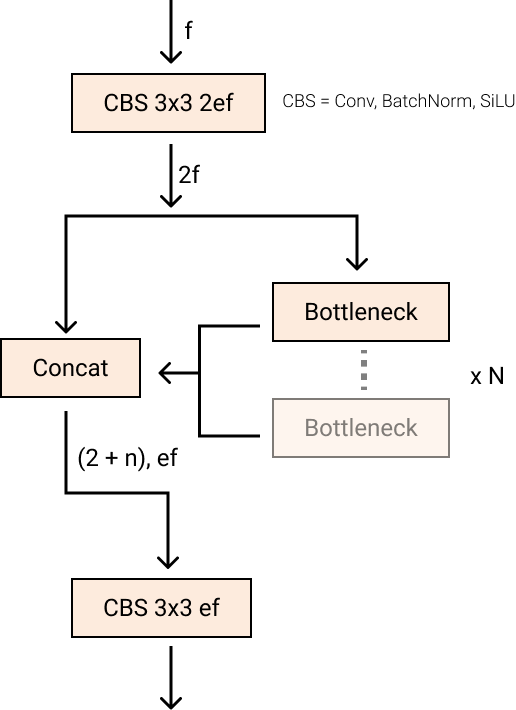}
    \caption{YOLO c2F module \cite{Roboflow}}
    \label{fig:c2f}
\end{figure}

YOLOv8 addresses several limitations of its predecessors, advancing both feature extraction and object detection performance through state-of-the-art backbone and neck architectures. Upon its debut on January 10th, 2023, YOLOv8 unveiled five iterations, spanning from YOLOv8n (the least extensive model, registering a 37.3 mAP score on COCO) to YOLOv8x (the most comprehensive model, marking a 53.9 mAP score on COCO) \cite{Roboflow}. It also reduced its parameter count by introducing a new C2f module (Figure-\ref{fig:c2f}). Unlike previous anchor-based approaches, YOLOv8 adopts an anchor-free split Ultralytics head, enhancing accuracy and detection efficiency. It strikes an optimized accuracy-speed trade-off, making real-time object detection across diverse applications more viable. Offering a variety of pre-trained models, YOLOv8 enables task-specific model selection, overcoming the generic model limitation. Enhanced training techniques contribute to better generalization and robustness across different datasets and real-world scenarios. Furthermore, the expanded class support broadens YOLOv8's applicability in object detection tasks. Technology optimizations in YOLOv8 have enhanced small object detection. We used three of its variants—nano (n) [3.2M parameters, 8.7B FLOPs], small (s) [11.2M parameters, 28.6B FLOPs], and medium (m) [25.9M parameters, 78.9B FLOPs]. Nano is the fastest, small balances speed and accuracy, while medium prioritizes accuracy over speed.

\subsection{License Plate Recognition with YOLOv8 (n,s,m)}
The YOLOv8 architecture was chosen for its reputation in real-time object detection and was employed on the LPR dataset. Three model variants of YOLOv8 namely, YOLOv8n, YOLOv8s, and YOLOv8m were evaluated under this study. The models were trained and tested on a fixed image size of 640 pixels. The table below provides a summary of the properties and performance metrics for each model variant.

\subsection{Character Recognition using YOLOv8s}
Following the LPR task, YOLOv8s model was selected and trained on the Character Recognition dataset due to its balanced trade-off between accuracy and computational efficiency as observed in the prior evaluations. The trained model was employed to detect characters which were subsequently cropped from the images.

\subsection{Character Sequencing}
In order to accurately sequence the detected characters from the License Plate Recognition (LPR) phase, a custom methodology was devised, leveraging the horizontal position of each character on the image plane. The set of detected characters is denoted as \( C = \{c_1, c_2, c_3, \ldots, c_n\} \), where each character \( c_i \) has an associated horizontal position \( x_i \) on the image. The primary objective was to derive a sequence \( S \) from the set \( C \) based on the ascending order of the horizontal positions. Initially, the set of horizontal positions, \( X = \{x_1, x_2, x_3, \ldots, x_n\} \), was meticulously sorted to obtain a sorted set \( X' = \{x_{i_1}, x_{i_2}, x_{i_3}, \ldots, x_{i_n}\} \) such that \( x_{i_1} < x_{i_2} < x_{i_3} < \ldots < x_{i_n} \). Following this, the sequence \( S = \{c_{i_1}, c_{i_2}, c_{i_3}, \ldots, c_{i_n}\} \) was constructed by mapping the sorted indices of \( X' \) back to the respective characters in \( C \). This methodology ensured a coherent sequencing of characters based on their x-axis positions, paving the way for accurate character recognition and subsequent license plate interpretation. It also significantly mitigates the likelihood of erroneous character sequencing, thus enhancing the overall performance of the license plate recognition system.

\subsection{Evaluation}
The performance of the models was rigorously evaluated using standard metrics such as Mean Average Precision (mAP), speed on both CPU and GPU, number of parameters, and Floating Point Operations Per Second (FLOPs). The results are comprehensively discussed in the subsequent section. The following evaluation metrics were used:

\subsubsection{Mean Average Precision (mAP)}
The mean Average Precision (mAP) is calculated using the formula:
\[
\text{mAP} = \frac{1}{N} \sum_{i=1}^{N} \text{AP}_{i}
\]
where \( \text{AP}_{i} \) is the average precision of the \(i\)-th class and \(N\) is the number of classes. mAP offers a single score to evaluate the detection performance across multiple classes.

\subsubsection{Loss Functions for Detection:}

\begin{enumerate}
    \item \textbf{Classification Loss:} 
The classification loss quantifies the divergence between the predicted class probabilities and the actual class labels, typically computed via Cross Entropy Loss. The formula for classification loss (\(L_{\text{cls}}\)) is given by:
\begin{equation}
    L_{\text{cls}} = -\sum_i y_i \log(p_i)
\end{equation}
where \(y_i\) is the ground truth label and \(p_i\) is the predicted probability for class \(i\).

    \item \textbf{Bounding Box Loss:}
The bounding box loss measures the error between the predicted bounding box coordinates and the ground truth coordinates. It is typically computed using Mean Squared Error (MSE). The formula for bounding box loss (\(L_{\text{bbox}}\)) is given by:
\begin{equation}
    L_{\text{bbox}} = \sum_i (x_i - \hat{x}_i)^2
\end{equation}
where \(x_i\) is the ground truth coordinate and \(\hat{x}_i\) is the predicted coordinate for dimension \(i\).

\end{enumerate}

The objective of these loss functions is to measure the discrepancy between the forecasted output and the genuine ground truth. A reduction in loss values signifies enhanced performance by the YOLOv8 model

\section{Results \& Discussion}
\begin{figure}[!htb]
    \centering
    \includegraphics[width=9cm]{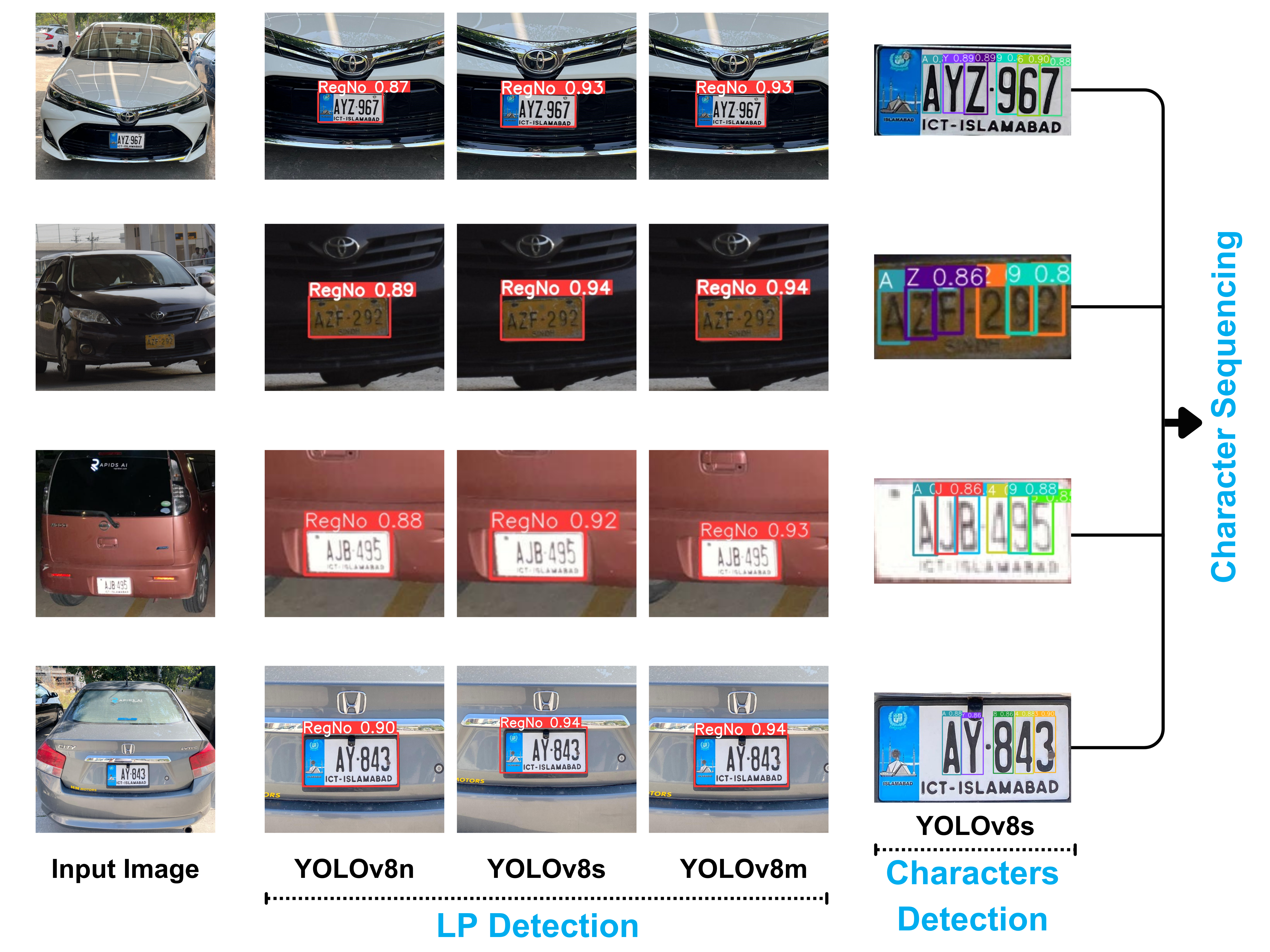}
    \caption{LPR and Character Recognition Results}
    \label{fig:results1}
\end{figure}

The computation for this study was carried out on an Intel(R) i7-12700k CPU operating at 3.61GHz, supported by 64 GB of RAM and a 24 GB Nvidia 3090 GPU. CUDA Version 12.0 facilitated GPU-accelerated computations. The evaluation test set included 225 images from the LPR Dataset and 3750 from the Character Recognition Dataset, offering a robust analysis of each model's proficiency. The training hyperparameters are delineated in Table-\ref{table:hyperparams}. A visual representation of the model performances is depicted in Figure-\ref{fig:results1} for a clearer understanding. Table-\ref{tab:performance_metrics_lpr} and \ref{tab:performance_metrics_chr} summarize the performance evaluation on the test set of LPR and Character Recognition Dataset. 
\begin{table*}[!htb]
\centering
\caption{Hyperparameters for training YOLOv8 models }
\label{table:hyperparams}
\begin{tabular}{|c|c|c|c|c|c|}
\hline
\textbf{Learning Rate} & \textbf{Batch Size} & \textbf{Epochs} & \textbf{Optimizer} & \textbf{Momentum} & \textbf{Weight Decay} \\
\hline
0.01 & 64& 500& AdamW & 0.937& 0.005 \\
\hline

\end{tabular}
\end{table*}

\begin{table}[!htb]
\centering
\caption{Performance Metrics for Yolov8 variants for License Plate Reg. No. (LPR) Test Data}
\resizebox{\columnwidth}{!}{%
\begin{tabular}{|c|c|c|c|c|c|}
\hline
\textbf{Yolov8 variants} & \textbf{Params (M)} & \textbf{Precision} & \textbf{Recall} & \textbf{mAP50} & \textbf{mAP50-95} \\
\hline
Nano & 3.2 &  0.964 & 0.876 & 0.918 & 0.669 \\
\hline
Small & 11.2 &  0.945 & 0.874 & 0.933 & 0.683 \\
\hline
Medium & 25.9 &  0.946 & 0.912 & 0.940 & 0.670 \\
\hline
\end{tabular}%
}
\label{tab:performance_metrics_lpr}
\end{table}

\begin{table}[!htb]
\centering
\caption{Performance Metrics for Yolov8 Character Test Data}
\begin{tabular}{|c|c|c|c|c|}
\hline
\textbf{Yolov8} & \textbf{Precision} & \textbf{Recall} & \textbf{mAP50} & \textbf{mAP50-95} \\
\hline
Small & 0.92 & 0.86 & 0.91 & 0.673 \\
\hline
\end{tabular}
\label{tab:performance_metrics_chr}
\end{table}

The evaluation of YOLOv8 variants on the LPR Test Data presents an insightful understanding of the performance trade-offs associated with different model sizes. Among the three variants, the YOLOv8 Medium variant exhibited the best performance with a mAP50 of 0.940, slightly higher than the Small variant which recorded a mAP50 of 0.933. The Medium variant also showed a superior Recall of 0.912, indicating its effectiveness in correctly identifying license plates amidst potential challenges such as occlusions or variations in license plate designs. However, it's noteworthy that the Nano variant, despite its smaller size, achieved the highest Precision of 0.964, which is indicative of its ability to deliver precise detections albeit at the expense of Recall (0.876).

On the other hand, the performance of YOLOv8 Small variant on Character Recognition dataset is commendable with a Precision of 0.92 and a Recall of 0.86, showcasing its proficiency in accurately recognizing characters. The mAP50 and mAP50-95 scores of 0.91 and 0.673 respectively, further corroborate the model's adeptness in character recognition tasks even under varying conditions, thereby demonstrating its applicability in real-world scenarios. The comparative analysis elucidates that while larger model variants like YOLOv8 Medium tend to have a better balance between Precision and Recall, smaller variants such as YOLOv8 Nano are more precise but less recall-oriented. This observation is pivotal for applications where either precision or recall is more critical, and an appropriate YOLOv8 variant can be chosen accordingly.
Moreover, the performance of YOLOv8 Small variant on Character Test Data underscores the versatility of the YOLOv8 architecture in handling both License Plate Recognition and Character Recognition tasks efficiently. The obtained results manifest a promising potential for employing YOLOv8 variants in real-world Intelligent Transportation Systems, especially where computational resources are a constraint, and a balance between speed and accuracy is sought.

\begin{figure}[!htb]
    \centering
    \includegraphics[width=8cm]{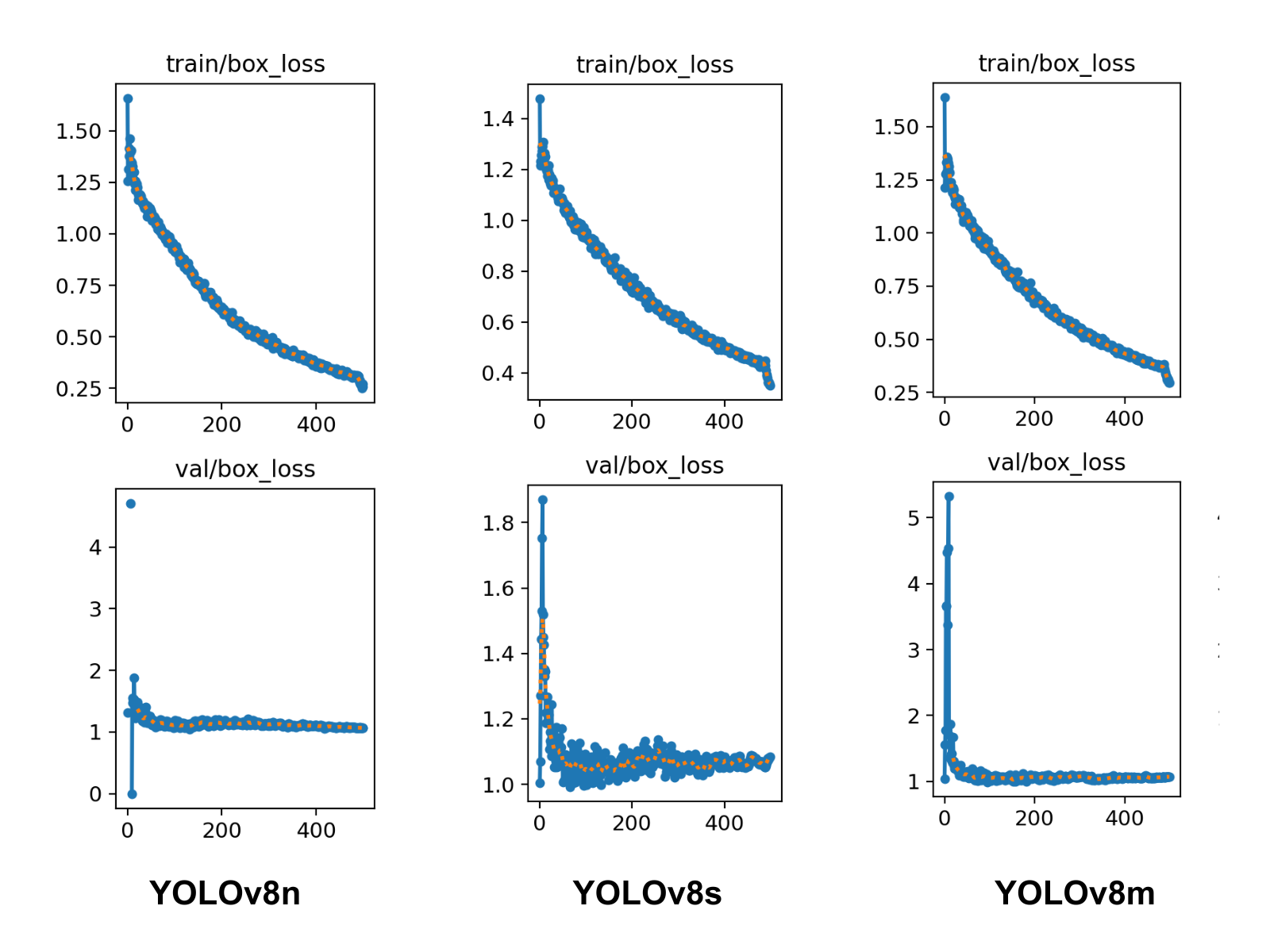}
    \caption{YOLO-V8 (n,s,m) - Training Performance on LPR Dataset}
    \label{fig:lpr}
\end{figure}

\begin{figure}[!htb]
    \centering
    \includegraphics[width=8cm]{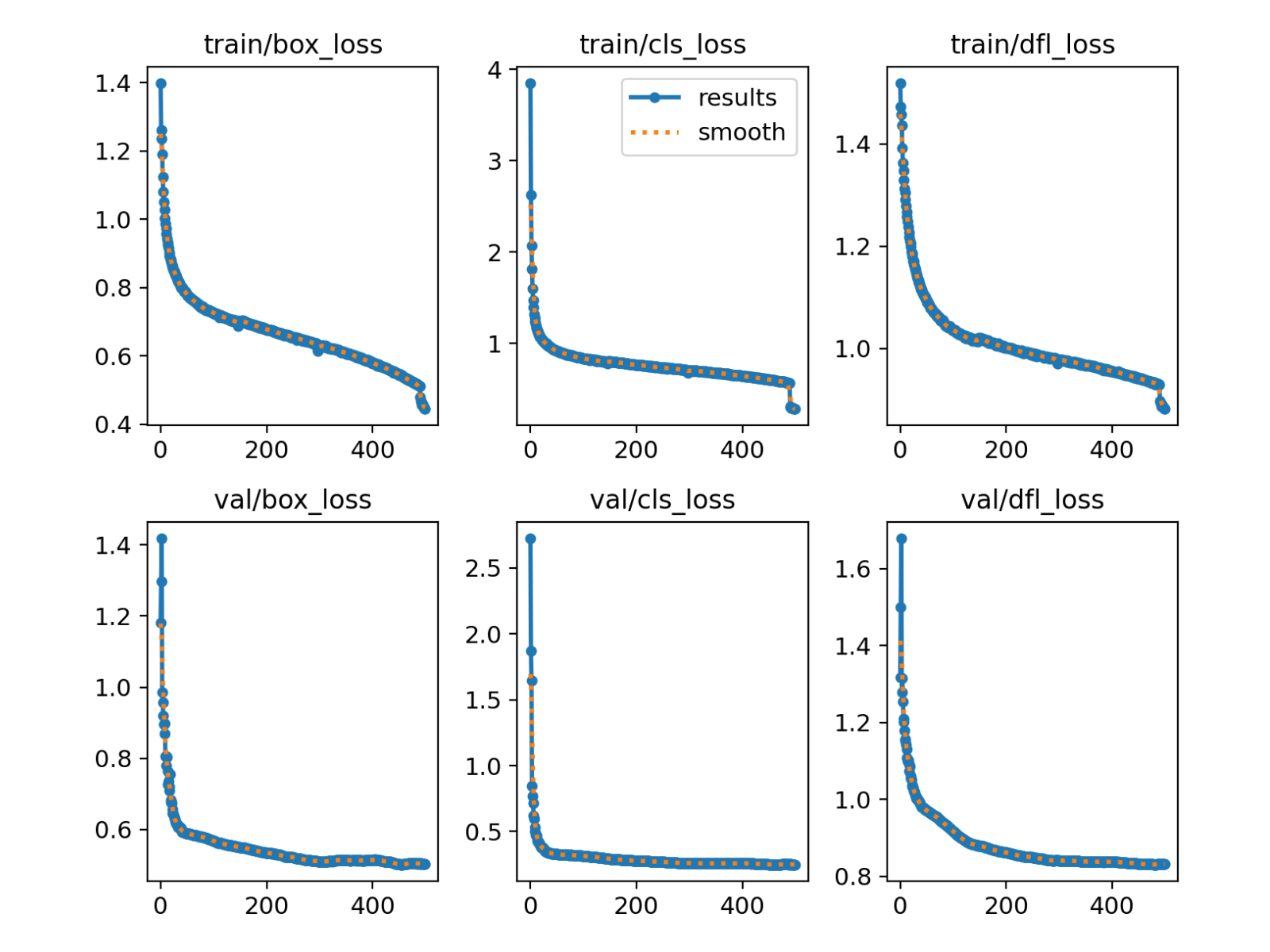}
    \caption{YOLO-V8s -  Training Performance on Character Recognition Dataset}
    \label{fig:chr}
\end{figure}

The exploration of YOLOv8 variants on LPR and Character Recognition tasks showcases YOLOv8 Nano's adeptness in LPR with a precision of 0.964, owing to its lightweight architecture of 3.2 million parameters. Conversely, YOLOv8 Small, with 11.2 million parameters, demonstrated a precision of 0.92 in Character Recognition, marking its suitability for this task. The union of YOLOv8 Nano for LPR and YOLOv8 Small for Character Recognition forms an optimized pipeline with a cumulative 14.4 million parameters, catering well to computational constraints on edge devices while preserving notable accuracy. These results pave the way for future real-world deployments on edge devices, aiming to prove this pipeline's real-time efficiency within Intelligent Transportation Systems, advancing towards more intelligent urban frameworks. 

\section{Conclusion}

In this study, the performance of three YOLOv8 models in License Plate and Character Recognition tasks was thoroughly examined. The YOLOv8s model emerged as a strong performer, registering a Precision of 0.945, Recall of 0.874, and mAP50-95 of 0.683 on the LPR test data, and a Precision of 0.92 on the Character Recognition test data. Additionally, the novel character sequencing method introduced in this study adeptly sequenced the detected characters with an outstanding accuracy of 99.8\%. The findings from this work underline the considerable potential of the YOLOv8 architecture in real-world LPR applications, especially when paired with innovative post-processing techniques like the character sequencing method proposed here. The curated datasets utilized in this study offer a significant resource for subsequent research in this domain.

Looking forward, future work may explore the performance of newer or alternative architectures, and the integration of semi-supervised or self-supervised learning methodologies to navigate scenarios with limited labeled data. Moreover, the lightweight nature of the YOLOv8 models, particularly YOLOv8(n and s), renders them suitable for deployment on edge devices, which is an avenue worth exploring for real-world, low-latency applications. Additionally, future work can address challenges from broken or damaged number plates, or those with paint off, to enhance license plate recognition robustness. Further refinement in character sequencing and extensive model evaluation across diverse real-world scenarios are crucial for advancing LPR and character recognition systems.

\bibliography{ref}
\end{document}